\documentclass[letterpaper, 10 pt, conference]{ieeeconf}  

\IEEEoverridecommandlockouts                              

\usepackage{graphics} 
\usepackage{epsfig} 
\usepackage{mathptmx} 
\usepackage{amsmath} 
\usepackage{amssymb}  
\usepackage{cite}
\usepackage{graphicx}
\usepackage{amsfonts} 
\usepackage{colortbl}
\usepackage{xcolor}
\usepackage{booktabs}
\usepackage{multirow}
\usepackage{float}
\usepackage{overpic}
\usepackage[normalem]{ulem}

\definecolor{colorfirst}{rgb}{.942,.855,.840} %
\definecolor{colorsecond}{rgb}{.780,.90,.980} %

\newcommand{\cellfirst}{\cellcolor{colorfirst}}
\newcommand{\cellsecond}{\cellcolor{colorsecond}}

\newcommand{\textfirst}{\colorbox{colorfirst}}
\newcommand{\secondtext}{\colorbox{colorsecond}}

\title{\LARGE \bf
ZeroDexGrasp: Zero-Shot Task-Oriented Dexterous Grasp Synthesis with Prompt-Based Multi-Stage Semantic Reasoning
}

\author{
\textbf{Juntao Jian$^{1}$}, \quad\quad \textbf{Yi-Lin Wei$^{2}$}, \quad\quad \textbf{Chengjie Mou$^{1}$}, \quad\quad \textbf{Yuhao Lin$^{2}$}, \quad\quad \textbf{Xing Zhu$^{3}$}, \\ 
\textbf{Yujun Shen$^{3}$},\quad\quad \textbf{Wei-Shi Zheng$^{2}$},\quad\quad \textbf{Ruizhen Hu$^{1\dagger}$}%
\thanks{$^{\dagger}$ Corresponding author.}%
\\[2mm]
{$^{1}$Shenzhen University \quad
$^{2}$Sun Yat-sen University \quad
$^{3}$Ant Group}
}

\begin{document}

\maketitle
\thispagestyle{empty}
\pagestyle{empty}


\begin{abstract}
Task-oriented dexterous grasping holds broad application prospects in robotic manipulation and human-object interaction. However, most existing methods still struggle to generalize across diverse objects and task instructions, as they heavily rely on costly labeled data to ensure task-specific semantic alignment. In this study, we propose \textbf{ZeroDexGrasp}, a zero-shot task-oriented dexterous grasp synthesis framework integrating Multimodal Large Language Models with grasp refinement to generate human-like grasp poses that are well aligned with specific task objectives and object affordances. Specifically, ZeroDexGrasp employs prompt-based multi-stage semantic reasoning to infer initial grasp configurations and object contact information from task and object semantics, then exploits contact-guided grasp optimization to refine these poses for physical feasibility and task alignment. Experimental results demonstrate that ZeroDexGrasp enables high-quality zero-shot dexterous grasping on diverse unseen object categories and complex task requirements, advancing toward more generalizable and intelligent robotic grasping.
\end{abstract}

\section{INTRODUCTION}

Task-oriented dexterous grasp synthesis has long been a challenging problem in the fields of robotic manipulation and human-object interaction. Unlike general dexterous grasp synthesis
primarily aims to ensure grasp stability or diversity, task-oriented dexterous grasping additionally requires producing multi-finger grasping poses that are well aligned with specific task objectives and object affordances (referred to as \emph{semantic requirements}).

To better adapt grasping hands to align semantic requirements, recent studies typically annotate grasp poses with type-specific labels~\cite{jian2023affordpose, yang2022oakink}, or further leverage large language models (LLMs) to generate diverse language instructions~\cite{wei2024graspGYS, li2024semgrasp, zhang2024nl2contact}. Both collect ``object-semantic-grasp" data pairs that are then used to train conditional generative models for grasp synthesis. Although these methods have achieved progress in handling given task instructions, these methods face major limitations: they require extensive manual data collection and cover only finite object categories, which severely restricts generalization to unseen objects and tasks in open-set scenarios.

\begin{figure}[!t]
   \includegraphics[width=0.98\linewidth]{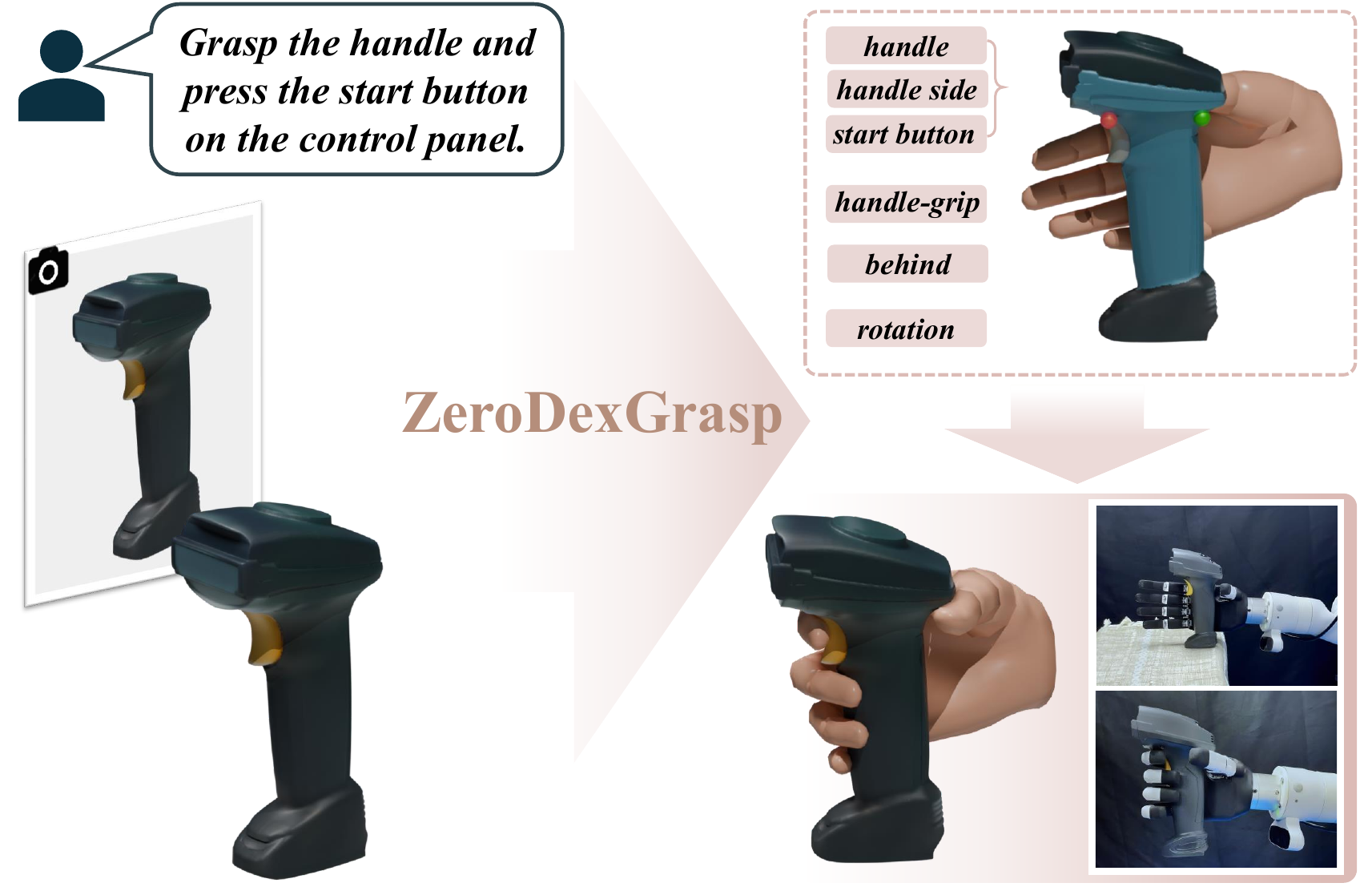}
   \caption{Given task instructions and target objects, \textbf{ZeroDexGrasp} synthesizes dexterous grasps in a zero-shot manner by reasoning contact information, hand position, orientation, grasp type, and followed by contact-guided refinement to ensure semantic alignment and physical feasibility.}
   \label{fig:teaser}
   \vspace{-18pt}
\end{figure} 

We aim to enhance the generalization of task-oriented dexterous grasping, particularly for open-set objects. This ambition introduces two primary challenges: (1) establishing reliable alignment between task instruction semantics and object affordances, and (2) bridging the significant conceptual gap from abstract, high-level semantic requirements to concrete, anthropomorphic dexterous grasp configurations.

We observe that in typical grasping behaviors, the hand's global properties, including final position, orientation, contact information, and overall grasp type, are primarily guided by semantic requirements, while the joints' configurations are more directly influenced by the object's local geometry. Besides, recent advancements in multimodal large language models (MLLMs) offer powerful capabilities in semantic understanding, spatial reasoning, and commonsense knowledge. These capabilities provide a feasible pathway for achieving zero-shot generalization of the hand’s global properties in dexterous grasping of diverse objects.

To this end, we propose \textbf{ZeroDexGrasp}, a zero-shot task-oriented dexterous grasp synthesis framework that integrates MLLMs with grasp optimization to achieve human-like grasp pose generation for open-set objects and task instructions, as shown in Fig.~\ref{fig:teaser}. ZeroDexGrasp comprises two key components, \emph{prompt-based multi-stage semantic reasoning} and contact-guided grasp refinement. The first component bridges the gap between task semantics, object affordances, and grasp poses by operating on discretized semantic representation, thereby producing an initial grasp pose and contact information that satisfies the semantic requirements. Specifically, part-based and point-based visual prompts are used to infer discrete object regions for hand and functional fingers contacts, textual candidate prompts, drawn from discrete sets, are used to infer the initial relative hand position and grasp type, and an imagination-based visual prompt method is used to guide the MLLMs to reason the hand's initial rotational pose. The second component leverages the inferred contact information to refine the initial grasp pose, synthesizing more reasonable, physically feasible, and task-aligned human-like dexterous grasps.

Our contributions are summarized as follows:
\begin{itemize}
\item We present ZeroDexGrasp, a zero-shot task-oriented dexterous grasp synthesis framework that integrates MLLMs with grasp optimization to generate high-quality dexterous grasps for open-set objects and tasks.
\item We develop a prompt-based multi-stage semantic reasoning based on discretized semantic representation to bridge task semantics, object affordances, and grasp poses, producing initial grasps and contact information that align with semantic requirements.
\item Extensive experiments on diverse open-set objects and tasks demonstrate the effectiveness and strong zero-shot generalization of ZeroDexGrasp.
\end{itemize}

\section{RELATED WORK}

\subsection{Task-Oriented Semantic Dexterous Grasping}
Task-oriented dexterous grasp synthesis aims to produce grasps that support downstream manipulation. Some methods abstract tasks as action labels and learn task-conditioned grasp distributions from annotated datasets~\cite{jian2023affordpose, yang2022oakink}. 

With the rapid progress of natural language processing and large language models, the diversity of task instructions has been enriched, and understanding unstructured task semantics has become feasible. Some works~\cite{ wei2024graspGYS, zhang2024nl2contact} encode task instructions and conditions a diffusion model on both text and object point clouds. SemGrasp~\cite{li2024semgrasp} finetunes pretrained MLLMs~\cite{lee2009advancesLLM}. Despite promising results, these approaches are constrained by limited training datasets and struggle to generalize to novel object categories. More recent works~\cite{zhong2025dexgraspvla, wei2025afforddexgrasp, zhang2025openhoi, jian2025g-dexgrap} have explored different strategies to enhance the generalization of dexterous grasp synthesis across unseen objects and diverse task instructions.  Dexgraspvla~\cite{zhong2025dexgraspvla} integrates large vision language models with grasp controllers to align grasp generation with task instructions. AffordDexGrasp~\cite{wei2025afforddexgrasp} introduce generalizable affordance representations, and G-dexgrasp~\cite{jian2025g-dexgrap} retrieves generalizable priors to bridge semantics and dexterous actions. However, these methods still demand expensive hand-specific data and training. DexVLG~\cite{he2025dexvlg} leverages geometry-based sampling and optimization to generate part-level grasp data, but focusing only on contact parts is insufficient to align high-level semantics requirements.

Different from the above methods, our approach achieves zero-shot alignment of task semantics, object affordances, and grasp poses, enabling more holistic semantics-to-grasp reasoning with strong generalization.

\subsection{Zero-Shot Robotic Grasping}
Zero-shot grasping synthesis refers to generating feasible grasps for diverse objects or tasks without requiring additional training or fine-tuning. Analytical approaches~\cite{ZapataImpata2019FastGC, wang2022dexgraspnet, wang2025unigrasptransformer} leverage object geometry to synthesize stable or varied grasps for diverse objects. In addition, with the development of large-scale grasping datasets, some works~\cite{fang2023anygrasp, Wu2024AnEF, Mousavian20196DOFGV} train foundation models for robotic grasping to improve performance on unseen objects. For low-degree-of-freedoms (Dofs) grippers, such as two-finger grippers, these models have been further adapted for task-oriented semantic grasping by leveraging Visual foundation Models~\cite{li2022grounded, sun2023going} to build semantic feature fields~\cite{kerr2023lerf, rashid2023language} or to identify task-relevant contact regions~\cite{Li2024ShapeGraspZT, Li2024SegGraspZT} in a zero-shot manner. Nevertheless, directly extending such strategies to dexterous hands remains challenging, as their high DoFs make it difficult to ensure reasonable, semantically meaningful global hand poses and physically plausible finger configurations.

\subsection{Visual Prompting in Multimodal Large Language Models}
MLLMs equip pre-trained LLMs with visual capabilities. While textual prompting~\cite{Gu2023ASS} in LLMs has been widely studied, visual prompting~\cite{Wu2024VisualPI} has emerged for more fine-grained and free-form visual instructions. Many studies leverage the annotation capabilities of VLMs to generate various information on raw images, such as bounding boxes~\cite{jiang2024joint} and keypoints~\cite{nasiriany2024pivot}. These annotations are then input as prompts to MLLMs, transforming complex visual tasks into more manageable forms~\cite{shtedritski2023does, yang2023dawn}. For instance, Shtedritski et al.~\cite{shtedritski2023does} found that VLMs like CLIP are capable of recognizing these specific visual annotations. SoM~\cite{yang2023setom} illustrated how object masks and numeric annotations can guide MLLMs to perform sophisticated visual reasoning tasks. Furthermore, visual prompting methods are being widely applied in robotic manipulation tasks~\cite{liu2024moka, huang2024rekep, huang2024copa} and have demonstrated excellent zero-shot generalization capabilities. This paper aims to explore the application of visual prompting techniques to dexterous hand grasping scenarios, and achieve zero-shot generalization.

\section{METHOD}

\begin{figure*}[!t]
   \includegraphics[width=1.0\linewidth]{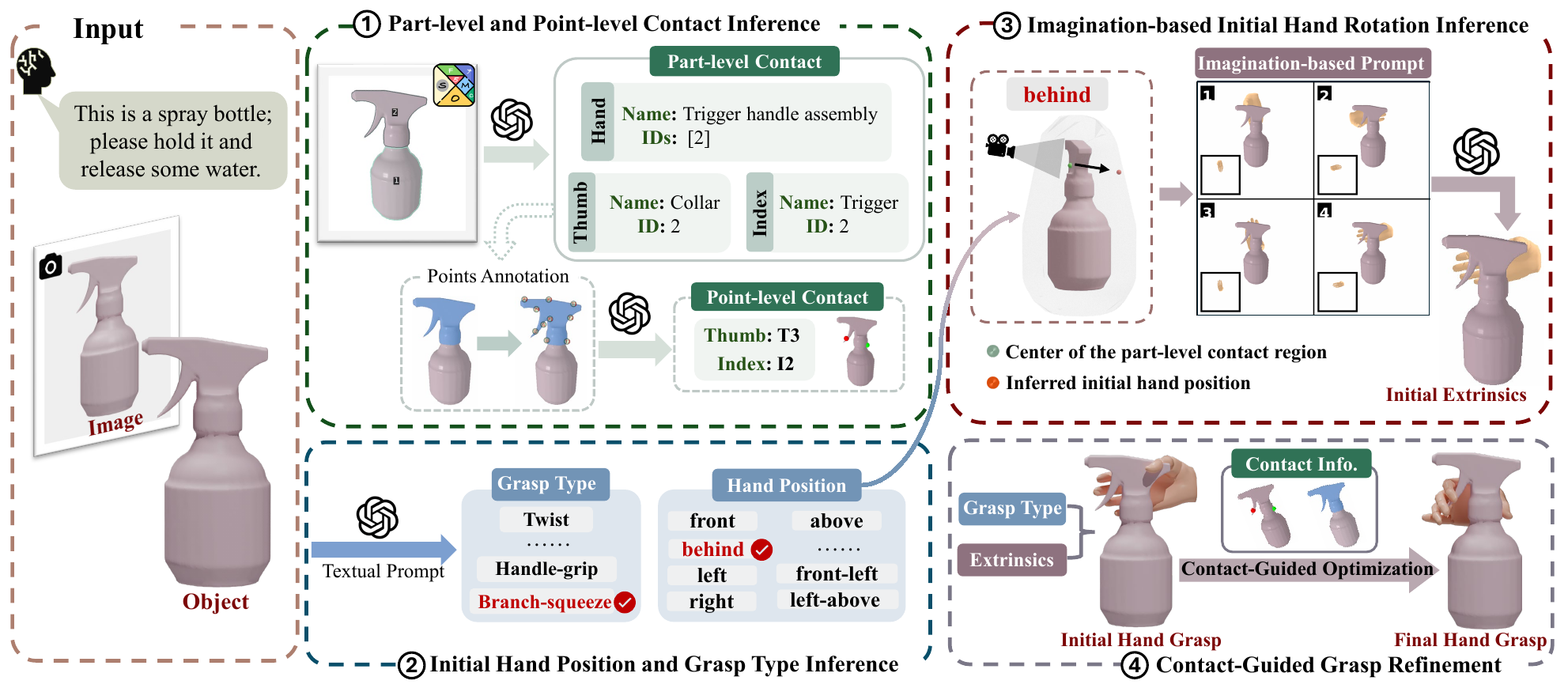}
   \caption{\textbf{Overall pipeline}. ZeroDexGrasp consists of two main components. The first is prompt-based multi-stage semantic reasoning, comprising three steps: (1) contact information inference, (2) grasp type and initial hand position inference, and (3) hand rotation inference. The second component, shown as part (4) in the figure, is contact-guided grasp optimization based on contact priors and the initial hand pose.}
   \label{fig:pipeline}
\end{figure*} 

Given a natural language task instruction $S$ and a target object $O$ as input, our goal is to synthesize grasp poses that are both physically feasible and semantically aligned. Specificly, model outputs a set of dexterous hand grasp parameters $G = \{T, R, \theta\}$, where $T \in \mathbb{R}^3$ and $R \in SO(3)$ represent the global translation and rotation of the hand (i.e., extrinsic parameters), and $\theta \in \mathbb{R}^{16}$ denotes the finger joint angles (i.e., intrinsic parameters). Furthermore, considering the crucial role of the thumb and index finger in both fine manipulation and stable grasping, we designate them as \emph{functional fingers} and assign them special emphasis during the inference process.

\noindent \textbf{Method Overview.} Our ZeroDexGrasp consists of two main components, as shown in Fig.~\ref{fig:pipeline}. The first component leverages MLLMs with prompt-based multi-stage semantic reasoning to perform zero-shot contact information and initial grasp pose reasoning: (1) inferring part-level and point-level contact region with part-based and point-based visual prompts; (2) predicting relative hand position and grasp type from textual candidate prompts of direction and grasp sets, producing semantically aligned initial hand position and joint configurations; and (3) determining the optimal initial hand rotation by introducing imagination-based visual prompts. The second component applies (4) contact-guided grasp refinement, ensuring effective hand-object contact and grasp stability. This process produces dexterous grasps that are both semantically aligned and physically feasible.

\subsection{Part-level and Point-level Contact Inference}
Humans instinctively determine specific contact regions by semantic requirements during object interaction, like grasping a knife by its handle or precisely positioning fingers on a spray bottle. To enable robots to replicate these commonsense behaviors, as illustrated in Fig.~\ref{fig:pipeline}, we employ powerful MLLMs (e.g., GPT-4.1) to infer 2D hand contact region $M_{hand}^{2d}$, functional fingers contact region $M_{func}^{2d}$ and contact keypoints $P_{func}^{2d}$ from the rendered object image $I$ and task $S$. Subsequently, we back-project these 2D predictions and apply feature classifying to obtain their corresponding 3D representations $\{M_{hand}^{3d}, M_{func}^{3d}, P_{func}^{3d}\}$.

\begin{figure}[!t]
   \includegraphics[width=0.98\linewidth]{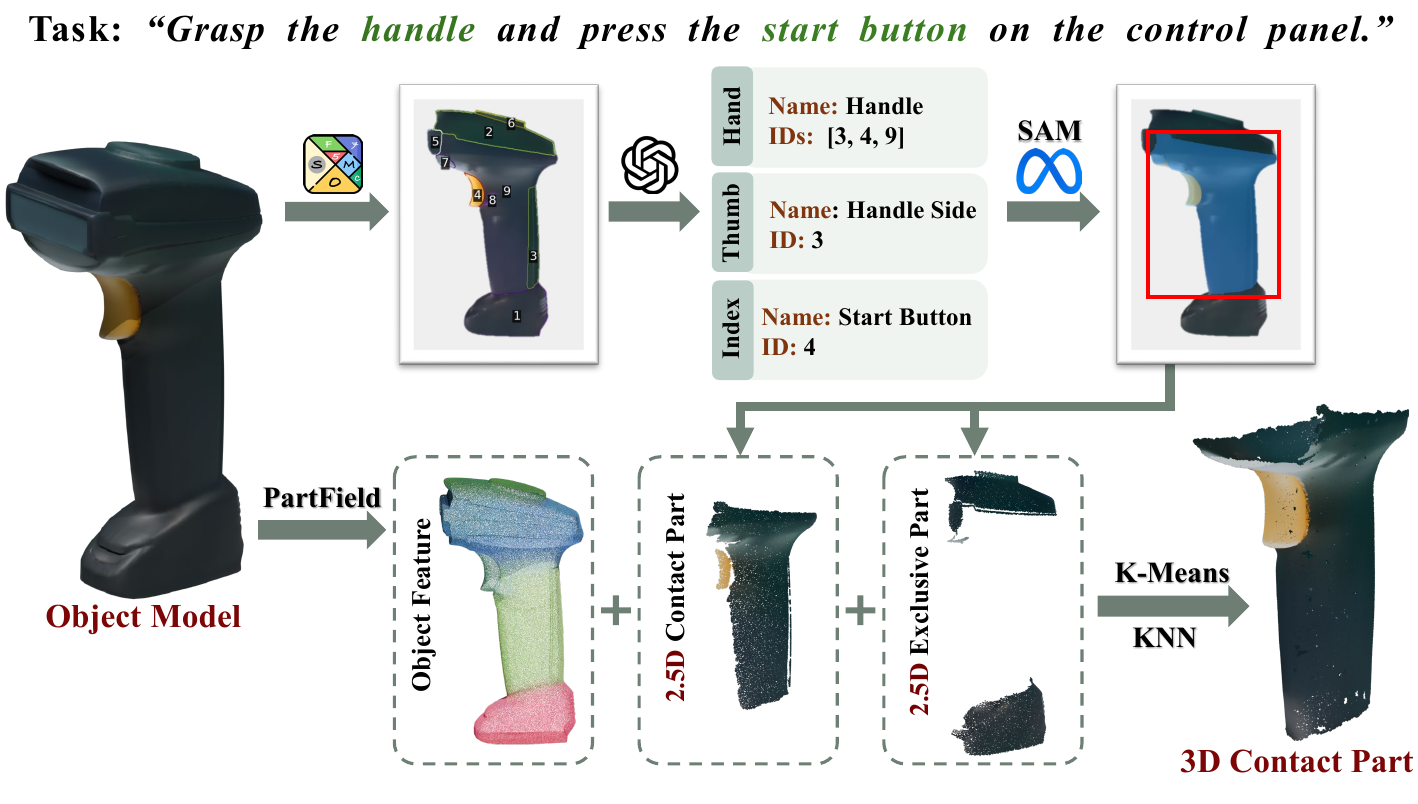}
   \caption{Pipeline of part-level contact inference. Semantic-aligned 2D part-level contact region are identified, back-projected to 2.5D, and the final 3D part-level contact region is inferred via feature clustering and classification.}
   \label{fig: contact-infer-image}
   \vspace{-20pt}
\end{figure}

\textbf{Part-level contact localization}  As shown in Fig.~\ref{fig: contact-infer-image}, we introduce an advanced visual prompting mechanism called Set-of-Mark (SoM)~\cite{yang2023setom}, which utilizes an advanced image segmentation model to automatically divide the input object image into discrete, distinct, and uniquely numbered regions. Based on this SoM-processed object image $I^{SoM}$, an MLLM separately infers the whole-hand contact masks and the functional finger contact masks by selecting annotated part IDs. Next, we merge the selected hand contact part masks and use the minimal bounding box enclosing them as a prompt for SAM~\cite{kirillov2023segany} to segment the complete hand contact part mask $M_{hand}^{2d}$. A similar process is used for the functional fingers contact parts to obtain a unified $M_{func}^{2d}$.

\textbf{Point-level contact localization} As illustrated in Fig.~\ref{fig:pipeline} by the spray bottle example, some tasks require more fine-grained contact information than part-level masks can provide to guide grasping, necessitating point-level contact reasoning for functional fingers. Specifically, we uniformly sample the contour of $M_{func}^{2d}$ to generate a set of discrete candidate contact points$\{ p_1, p_2, ...,p_n \}$. 
This significantly enhances the precision with which language instructions can be aligned and localized within the visual space. Each candidate point is then assigned a unique ID and visually marked on the image, resulting in a prompted image $I^{points}$, which serves as a visual prompt to infer the specific 2D contact points, denoted as $P_{func}^{2d} = \{P_{thumb}^{2d}, P_{index}^{2d}\}$, where $P_{thumb}^{2d}$ and $P_{index}^{2d}$ correspond to the thumb and index finger respectively.  Note that point-level localization is not always required, as part-level contact information typically suffices for most grasping tasks. Therefore, for efficiency, point-level contact inference is applied conditionally in deployment. Criteria for omitting inference include concurrent MLLM evaluation or when the functional fingers share the same inferred contact part information (ID and name).

\textbf{3D Contact Representation Extraction}
Given the camera parameters, we back-project the 2D information into 3D space extracting single-view representations: $\{M_{hand}^{2d}, P_{func}^{2d}\} \rightarrow \{M_{hand}^{2.5d}, P_{func}^{2.5d}\}$. For simplify, we assume $M_{hand}^{2d} = M_{func}^{2d}$ due to their typical similarity. 

The transformation $P_{func}^{2.5d} \rightarrow P_{func}^{3d}$ is achieved by expanding $P_{func}^{2.5d}$ into a local spherical region.
For $M_{hand}^{2.5d} \rightarrow M_{hand}^{3d}$, We adopt a feature-based classification strategy. Specifically, object pixels outside $M_{hand}^{2d}$ are labeled as \emph{exclusive pixels} and back-projected to obtain $M_{ex}^{2.5d}$, forming a binary classification with $M_{hand}^{2.5d}$. Then, we employ PartField~\cite{liu2025partfield} to extract point-level semantic features for the full object $O$. The unlabeled points near $M_{hand}^{2.5d}$ are first grouped by K-Means, and each cluster is assigned a label predicted by a KNN classifier trained on the known $M_{hand}^{2.5d}$ and $M_{ex}^{2.5d}$ point features, with all points in the cluster inheriting that label.

\subsection{Initial Hand Position and Grasp Type Inference}
When interacting with objects, some hand configurations are clearly unsuitable. For example, placing the palm above a phone screen or using a knife-handle grasp pose for holding a phone. To ensure task-consistent initial hand placement and grasp selection, we employ MLLMs with textual candidate prompts to infer the initial relative hand position $T'$ and grasp type $G$.

\textbf{Relative Hand Position} We define a discrete set of candidate hand approach directions $D_{text}$, including six cardinal direction set {``front", ``behind", ``left", ``right", ``above", ``below"} and 12 diagonal directions formed by pairwise combinations. A local coordinate frame is established at the center $C$ of $M_{hand}^{3d}$. 
The ``front" direction is deineded as the unit vector pointing from the center $C$ toward the camera, denoted $\vec{D}_{\text{front}}$, with the opposite direction $\vec{D}_{\text{behind}} = -\vec{D}_{\text{front}}$. The remaining cardinal directions are defined as:
\[
\begin{aligned}
\vec{D}_{\text{above}} &= \hat{y}, & \vec{D}_{\text{below}} &= -\vec{D}_{\text{above}}, \\
\vec{D}_{\text{right}} &= \vec{D}_{\text{front}} \times \vec{D}_{\text{above}}, & \vec{D}_{\text{left}} &= -\vec{D}_{\text{right}}.
\end{aligned}
\]
where $\hat{y}$ is the upward direction in the world frame. Diagonal directions are then obtained by normalizing the sum of two base vectors, e.g.,
\[
\vec{D}_{\text{front-right}} = \frac{\vec{D}_{\text{front}} + \vec{D}_{\text{right}}}{\|\vec{D}_{\text{front}} + \vec{D}_{\text{right}}\|}.
\]

Based on the textual candidate prompt, an MLLM selects the most semantically aligned textual direction $D_{text}^*$ from $D_{text}$ and maps to obtain its direction vector $\vec{D}^*$.  
The initial 3D hand position $T'$ can is computed by casting a ray from the center $C$ along $\vec{D}^*$ and finding its intersection with the object's expanded convex hull $H_{obj}$:
\begin{equation}
T' = \text{Intersect}(\text{ray}(C, \vec{D}^*), H_{obj}).
\end{equation}

\textbf{Grasp Type} To ensure that the initial finger configuration aligns with the semantic requirements of the object, we predefine a set of common grasp types $G_{set}$ following~\cite{jian2025g-dexgrap}. The MLLM selects the most appropriate grasp type $G \in G_{set}$, and its associated parameters $\theta'$ to serve as the initial fingers configurations.

\subsection{Imagination-based Initial Hand Rotation Inference}
A dexterous grasp's success and semantic alignment also rely on reasonable hand rotation, defined by the palm direction $\vec{D}_{palm}$ and the finger direction $\vec{D}_{finger}$. Specifically, $\vec{D}_{palm}$ is the inward normal of the palm surface, and $\vec{D}_{finger}$ denotes the overall direction of finger extension when the hand is naturally stretched, shown in Fig.~\ref{fig:filter} (a). 

As a primary constraint, we define the optimal palm orientation $\vec{D}_{palm}^{*}$ to oppose the contact-related direction obtained in the previous subsection, i.e.,
\begin{equation}
    \vec{D}_{palm}^{*} = - \vec{D}^*.
\end{equation}
For the optimal finger direction $\vec{D}_{finger}^{*}$, we propose an \emph{``imagination''-based} visual prompt method. A set of candidate hand states is generated by rotating the hand around $\vec{D}_{palm}^{*}$, and the MLLM evaluates the rendered candidate poses to select the optimal rotational component $R'$.  

Specifically, the process is conducted in a simulation environment containing the hand model and the target object $O$. By placing the hand at the inferred initial position $T'$ with the original joint configuration $\theta_0$, and rotating around $\vec{D}_{palm}^{*}$, we generate $K$ candidate hand states: $\{ (\theta_0, T', R_k) \}_{k=1}^K$, where each $R_k$ represents a hand rotation. 

For each candidate, two images are rendered: one with the hand and object ($I^{h,o}_k$) and one with the hand only ($I^{h}_k$). These are then fused into a composite view $I_k$. Collecting and labeling all such composites produces a comprehensive visual prompt $I^{imag}$, which is fed into an MLLM. The MLLM evaluates the set and selects the optimal rotation $R'$.

Through these three sub-processes of prompt-based multi-stage semantic reasoning, our framework zero-shot infers an initial grasp pose $G' = (\theta', T', R')$ and contact information $\{ M_{hand}^{3d}, M_{func}^{3d}, P_{func}^{3d} \}$ strongly correlated with semantics requirements.

\subsection{Geometry-Guided Verification}
To enhance the physical plausibility of semantic reasoning results, we apply an additional geometry-guided verification, including imagined hand rotation filtering and point-level contact validation.

\begin{figure}[!t]
   \includegraphics[width=0.98\linewidth]{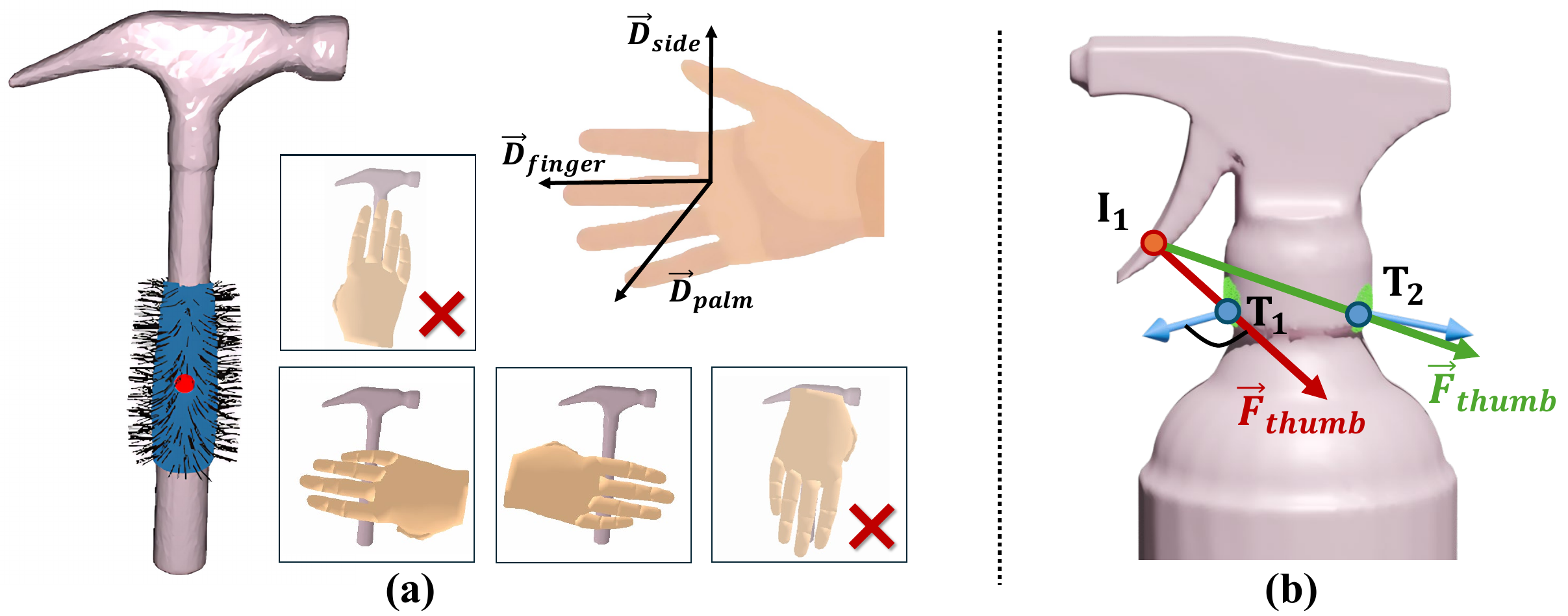}
   \caption{Illustration of eometry-Guided Verification. (a) Hand rotation filtering via local surface normals, where the red point denotes the nearest surface point and the blue points indicate its neighbors. (b) point-level contact validation using force-normal consistency.}
   \label{fig:filter}
   \vspace{-18pt}
\end{figure}

\textbf{Imagined Hand Rotation Filtering} During the initial hand rotation inference stage, when point-level contact inference is not required, a candidate hand state $(\theta_0, T', R_k)$ may be filtered based on alignment between functional finger directions and local surface normals. Specifically, we identify the nearest surface point $\mathbf{p}^*$ on the contact region relative to the hand: $\mathbf{p}^* = \text{Intersect}(\text{ray}(C, \vec{D}^*), O)$, and sample its neighboring points to obtain normals $\{\mathbf{n}_i\}$. A candidate state is filtered before the MLLM inference if its finger direction $\vec{D}_{finger}$ does not satisfy the angular alignment constraint with any local normal, i.e.,
\begin{equation}
\forall i \quad \text{s.t.} \quad
\frac{|\mathbf{n}_i \cdot \vec{D}_{finger}|}{\|\mathbf{n}_i\| \, \|\vec{D}_{finger}\|} < \tau_n,
\end{equation}
where $\tau_n$ is a predefined threshold. This ensures consistency between the finger orientation and local object geometry. As shown in Fig.~\ref{fig:filter} (a), candidate rotations whose $\vec{D}_{finger}$ aligns with the hammer handle's principal axis are filtered out, since the sampled surface points lack normals parallel to $\vec{D}_{finger}$.

\textbf{Point-level contact Validation}  For point-level contact inference results, we validate them using force-normal consistency. The result that do not meet the validation criteria will be re-inferred. Specifically, using the back-projected $P_{func}^{2.5d}= \{P_{thumb}^{2.5d},P_{index}^{2.5d} \}$ define a local force direction for each functional finger:
\begin{equation}
\vec{F}_{thumb} = P_{thumb}^{2.5d} - P_{index}^{2.5d}, \quad
\vec{F}_{index} = -\vec{F}_{thumb}.
\end{equation}

For each contact point $P_{k}^{2.5d}$, we sample neighboring points $\{\mathbf{q}_i\}$ with normals $\{\mathbf{n}_i\}$ from $P_{k}^{3d}$. The referred point-level contact is discarded if no normal is approximately aligned with the force direction, i.e.,
\begin{equation}
\forall i \quad \text{s.t.} \quad  
\frac{\mathbf{n}_i \cdot \vec{F}_k}{\|\mathbf{n}_i\| \, \|\vec{F}_k\|} < \tau_m.
\end{equation}

Fig.~\ref{fig:filter} (b) illustrates an example. When $P_{index}^{2.5d} = I_1$, assigning $P_{thumb}^{2.5d} = T_1$ violates the force-normal consistency, as the angle between $\vec{F}_{thumb}$ and the normals of all sampled points is too large to satisfy the above constraint. In contrast, selecting $P_{thumb}^{2.5d} = T_2$ satisfies the condition.

\subsection{Contact-Guided Grasp Refinement}

Given the infered initial grasp $G' = (\theta', T', R')$, the hand mesh can be extracted by employing the differentiable kinematics layer. To ensure final dexterous grasps that are both semantically aligned and physically feasible, we further optimize $G'$ based on inferred contact information $\{ M_{hand}^{3d}, P_{func}^{3d} \}$, where $P_{func}^{3d}=M_{func}^{3d}$ if point-level contact localization is not required. 

\begin{table*}
\centering
\caption{Quantitative comparison on the test set with diverse object categories. GraspTTA$^\dag$ and AffordPose$^\dag$ use CLIP to encode task instructions. Best and second-best results are marked in \textfirst{\textbf{red}} and \secondtext{\textbf{blue}}, P.Score via MLLM/User.}
\resizebox{0.95\textwidth}{!}{
\begin{tabular}{l|p{1.5cm}<{\centering}p{1.5cm}<{\centering}p{1.5cm}<{\centering}p{1.6cm}<{\centering}|p{1.6cm}<{\centering}p{1.5cm}<{\centering}p{1.5cm}<{\centering}}
\hline
\multirow{2}{*}{Experiments} & \multicolumn{4}{c|}{Quality and Stability}                                                                         & \multicolumn{3}{c}{Semantic Alignment}                                       \\ 
\cline{2-8}
                             & P.Vol.($cm^3$)$\downarrow$       & P.Dep.($cm$)$\downarrow$     & Sim.Dis.($cm$)$\downarrow$     & Sim.Suc.$(\%)$    & Part Acc.$(\%)$      & S.C.Ratio$(\%)$     & P.Score$(\%)$    \\ 
\hline
GraspTTA$^\dag$~\cite{jiang2021handTTA}      & 10.847               & 1.672                  &  4.916             & 37.458            & 47.886               & 43.896              & 2.05/1.26   \\
AffordPose$^\dag$~\cite{jian2023affordpose}   & 7.846               & 1.437                  & \cellsecond 3.919   & 54.862            & 53.172               & 50.674              & 3.12/2.21   \\
DexGYS~\cite{wei2024graspGYS}                 & 10.489               & 1.686                  & \cellfirst 3.507     & 42.288            & 38.729               & 38.058              & 2.96/1.95   \\
Ours$^*$                                & \cellfirst 2.489      & \cellfirst 0.315           & 4.936                 & \cellsecond 60.146     & \cellsecond 86.970   & \cellsecond 81.929   & \cellsecond 4.01/3.83   \\
Ours                                    &\cellsecond 3.322      &\cellsecond 0.358           & 4.550                 & \cellfirst 63.101      & \cellfirst 90.075    & \cellfirst 83.831    & \cellfirst 4.00/3.85 \\
\hline
\end{tabular}
}
\label{tab:comparison_results}
\end{table*}

Specifically, we define a set of optimization energy terms to refine the grasp. Firstly, a functional finger contact energy $E_{cont}^{fun}$ encourages the functional fingers to reach the inferred contact keypoints $P_{func}^{3D}$, and a non-functional finger contact energy $E_{cont}^{unf}$ enhances overall contact consistency. Both terms are weighted by the hand contact probability map $C$ estimated from prior methods~\cite{jian2025g-dexgrap, jiang2021handTTA}:
\begin{equation}
E_{cont}^{fun} = \frac{1}{n_{func}} \sum_{j=1}^{n_{func}} C_j \min_{i\in N_{fun}} \| V_j - P^{3d}_{func_{i}} \|,
\end{equation}
\begin{equation}
E_{cont}^{unf} = \frac{1}{n_{unf}} \sum_{t=1}^{n_{unf}} C_t \min_{i\in N_{part}} \| V_t - M^{3d}_{hand_i} \|,
\end{equation}
$C_j$ denotes the contact probability of the $j$-th functional fingers vertices $V_j$, $n_{func}$ is the total number of functional finger vertices, $P_{func_i}^{3d}$ denotes the $i$-th point-level contact point, and $N_{func}$ is the total number of point-level contact points. Similar representations are used for $E_{cont}^{unf}$.

Additionally, similar to~\cite{jian2025g-dexgrap}, we incorporate a hand contact map energy $E_{cmap}^{hand}$, a penetration energy $E_{pen}$, a self-penetration energy $E_{spen}$, a force-closure energy $E_{fc}$, and a joint limit energy $E_{pip}$. The final optimization minimizes a weighted sum of these energy terms:
\begin{equation}
\begin{split}
E = &\; \lambda_1 E_{cont}^{func} + \lambda_2 E_{cont}^{unf} + \lambda_3 E_{cmap}^{hand} 
\\
&+ \lambda_4 E_{pen} + \lambda_5 E_{spen} + \lambda_6 E_{fc} + \lambda_7 E_{pip}
\end{split}
\end{equation}
where $\lambda_1$–$\lambda_7$ denote the coefficients for the respective loss terms.

\section{EXPERIMENTS}

\subsection{Experimental Setup}
\textbf{Dataset}
We use the AffordPose dataset~\cite{jian2023affordpose}, which contains over 20k grasps across 11 grasp-related object categories and 5 grasp types, to train our comparison methods, and additionally generate task instructions as input. 

For testing, there is no publicly available off-the-shelf benchmark with sufficiently diverse objects and annotated task instructions for task-oriented dexterous grasping. Therefore, we follow~\cite {jian2025g-dexgrap} to construct a test set comprising more than 50 object categories with well-defined grasp semantics (including the 11 categories from AffordPose), 200 object instances, and 280 task instructions to evaluate the zero-shot generalization ability of our method.  This test dataset includes common household categories such as ``coffee machines", ``teapots", ``buckets", and ``suitcases", ensuring diversity in task semantics, object affordances, and geometric shapes. Besides, each test sample is annotated with task-aligned part-level contact regions, providing a reliable basis for task-conditioned grasp evaluation. 

\textbf{Evaluation Metrics}
We adopt commonly-used metrics from~\cite{jiang2021handTTA, jian2023affordpose, jian2025g-dexgrap} to assess the plausibility and task alignment of the generated grasps. For plausibility, we measure \textbf{Penetration Depth} (P.Dep., $cm$) and \textbf{Penetration Volume} (P.Vol., $cm^3$), which quantify the overlap between the object and hand meshes. \textbf{Simulation Displacement} (Sim.Dis., $cm$) measures the displacement of the object over time when subjected to the generated grasps. \textbf{Simulation Success} (Sim.Suc., \%) is the ratio of grasps that maintain continuous contact with penetration depth below $1 cm$. For simulation-based metrics, we exclude tasks involving ``Twist'' to avoid confounding effects.
For semantic alignment, we use the \textbf{Part Accuracy} (Part Acc. \%) metric to verify whether the generated hand makes contact with the object at the ground-truth contact parts. 
Second, we introduce the \textbf{Semantic Contact Ratio} (S.C.Ratio, \%), which calculates the proportion of finger vertices that are in contact with the correct semantic part of the object. 
Finally, we randomly sampled grasps to assess the \textbf{Perceptual Score} (P.Score 1-5) via MLLM scoring (one result per object category) and a user study with 15 participants (10 results).

\textbf{Implementation Details}
For prompt-based multi-stage semantic reasoning, we employ the Gemini-2.5 series to infer hand position, while the GPT-4.1/o4 series handles other semantic reasoning tasks. We adopt the five common grasp types defined in~\cite{jian2025g-dexgrap}. These grasp types cover the most common grasping tasks and are easy to extend with an arbitrary number of additional grasp demonstrations. Besides, we render a candidate imagined image $I_k$ every $90^\circ$ ($K=4$) around the palm direction. In the geometry-guided verification, we set $\tau_n=0.85$ and  $\tau_m=0.7$.

During the optimization stage,  the object's surface is represented by a uniformly sampled 4096 points with normals as the input. We employ the Adam optimizer for 600 iterations with an initial learning rate of 0.005 and an exponential decay factor of 0.98 every 10 epochs. The weights of the objective function are set as follows: $\lambda_1=60$, $\lambda_2=30$, $\lambda_3=50$, $\lambda_4=20$, $\lambda_5=1$, $\lambda_6=1$, $\lambda_7=1$.

\subsection{Comparison}

\begin{table*}
\centering
\caption{Quantitative evaluation of our ablation study to validate the contribution of our design. Best and second-best results are marked in \textfirst{\textbf{red}}, P.Score via MLLM/User.}
\resizebox{0.95\textwidth}{!}{
\begin{tabular}{l|p{1.5cm}<{\centering}p{1.5cm}<{\centering}p{1.5cm}<{\centering}p{1.6cm}<{\centering}|p{1.6cm}<{\centering}p{1.5cm}<{\centering}p{1.5cm}<{\centering}}
\hline
\multirow{2}{*}{Experiments} & \multicolumn{4}{c|}{Quality and Stability}                                                                         & \multicolumn{3}{c}{Semantic Alignment}                                       \\ 
\cline{2-8}
                    & P.Vol.($cm^3$)$\downarrow$       & P.Dep.($cm$)$\downarrow$     & Sim.Dis.($cm$)$\downarrow$     & Sim.Suc.$(\%)$    & Part Acc.$(\%)$      & S.C.Ratio$(\%)$     & P.Score$(\%)$    \\ 
\hline
W/O Contact         & 4.767             & 0.417                & \cellfirst 4.007    & \cellfirst 64.421         & 53.518            & 54.598              & 3.48/2.80   \\
W/O Initial        & 4.569             & 0.440                & 5.595                & 52.015        & 79.226            & 73.114           & 2.69/1.87   \\
W/O Refine          & 11.443            & 1.585                & 4.301               & 43.135         & 73.940            & 71.963           & 3.55/2.75   \\
Ours            &\cellfirst 3.322   &\cellfirst 0.358         & 4.550                & 63.101        & \cellfirst 90.075  & \cellfirst 83.831   & \cellfirst 4.00/3.85 \\
\hline
\end{tabular}
}
\label{tab: ablation_results}
\end{table*}
\begin{figure}[!t]
   \includegraphics[width=1.0\linewidth]{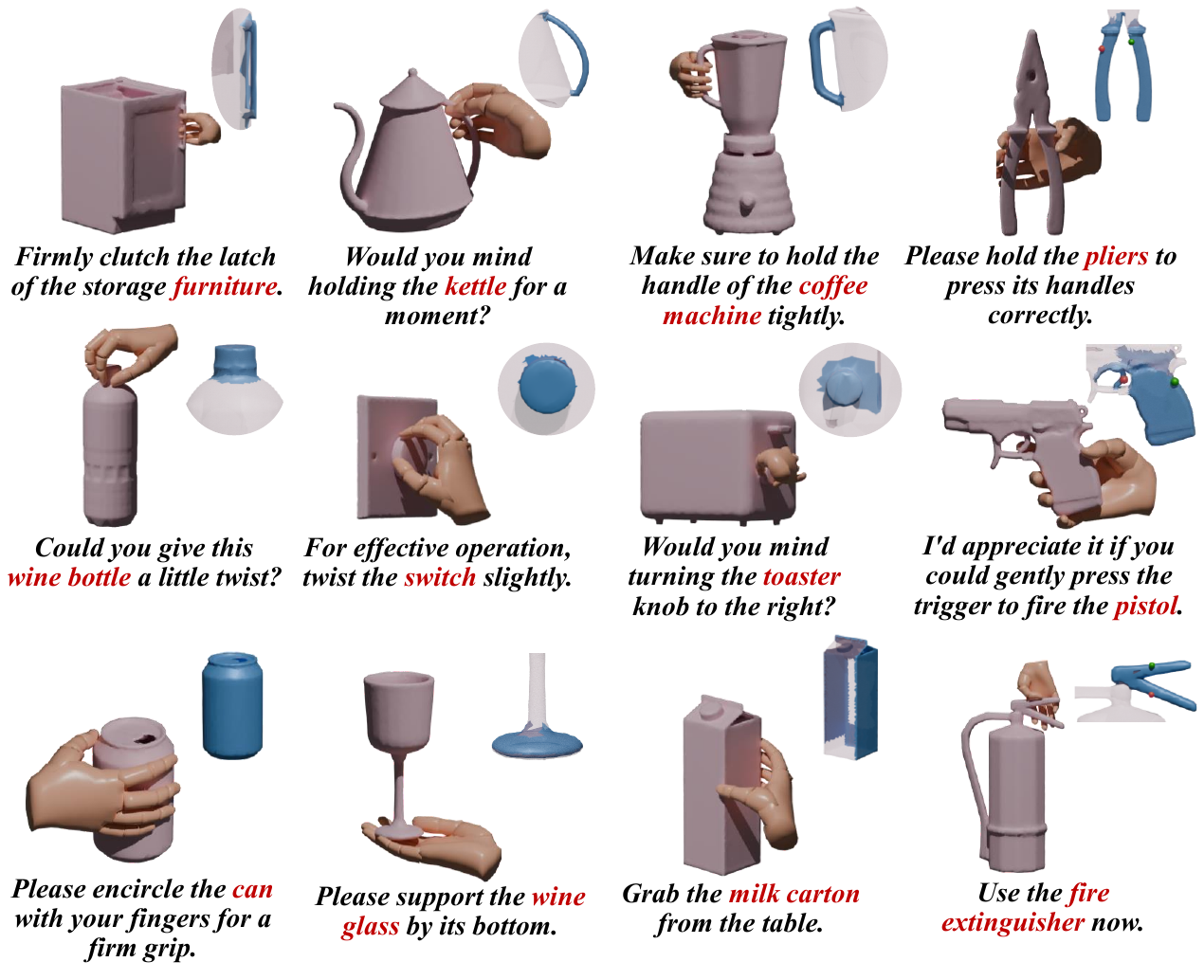}
   \caption{Qualitative results of \emph{Ours}. ZeroDexGrasp achieves zero-shot, semantically aligned, and physically feasible grasps for diverse objects and tasks. The top-right corner shows the inferred part-level contact region (blue), point-level contact for the index finger (red) and the thumb (green).}
   \label{fig:comparison-results}
   \vspace{-18pt}
\end{figure} 

\textbf{Baselines}
We compare our method with open-source approaches of task-oriented dexterous grasp synthesis. Firstly, we adapt GraspTTA~\cite{jiang2021handTTA}, the general-purpose dexterous grasp synthesis method by encoding task instructions with a pre-trained CLIP model~\cite{radford2021learning} and incorporating these features as conditional inputs, denoted as GraspTTA$^\dagger$. In addition, existing approaches generally rely on action labels or textual task embeddings. For the former, we select AffordPose~\cite{jian2023affordpose} as a representative baseline and extend it to process language instructions by replacing its original affordance-label input with CLIP-encoded textual features. Finally, we include an open-source diffusion-based method for task-oriented dexterous grasp synthesis, DexGYS~\cite{wei2024graspGYS}, as an additional baseline. We also provide two variants of our approach. \emph{Ours*} uses an MLLM to select the best view from four surrounding rendered images of the object, while \emph{Ours} selects the view manually.

We report the quantitative evaluations in Tab.~\ref{tab:comparison_results}. The performances of \emph{Ours*} and \emph{Ours} are highly similar. The former selects the best render viewpoint based on an MLLM, which lead to slightly worse results since it introduces some errors. Nevertheless, they consistently outperform existing approaches. Specifically, we achieve reductions by more than $50\%$ on penetration volume and depth. Although our method is not SOTA in simulation displacement, this result may be beneficial for high interpenetration. In contrast, we maintain competitive stability while ensuring minimal penetration, suggesting that the resulting grasps are physically more reliable. Moreover, our method demonstrates strong semantic alignment, nearly doubling that of prior works. This improvement can be attributed to the superior generalization ability of our framework, while baseline methods struggle to transfer across unseen object categories and task instructions. As shown in Fig.~\ref{fig:comparison-results}, our approach could synthesize semantically aligned and physically feasible grasp poses for diverse objects and task instructions in a zero-shot manner, even when there are significant differences.

\begin{figure}[!t]
   \includegraphics[width=0.98\linewidth]{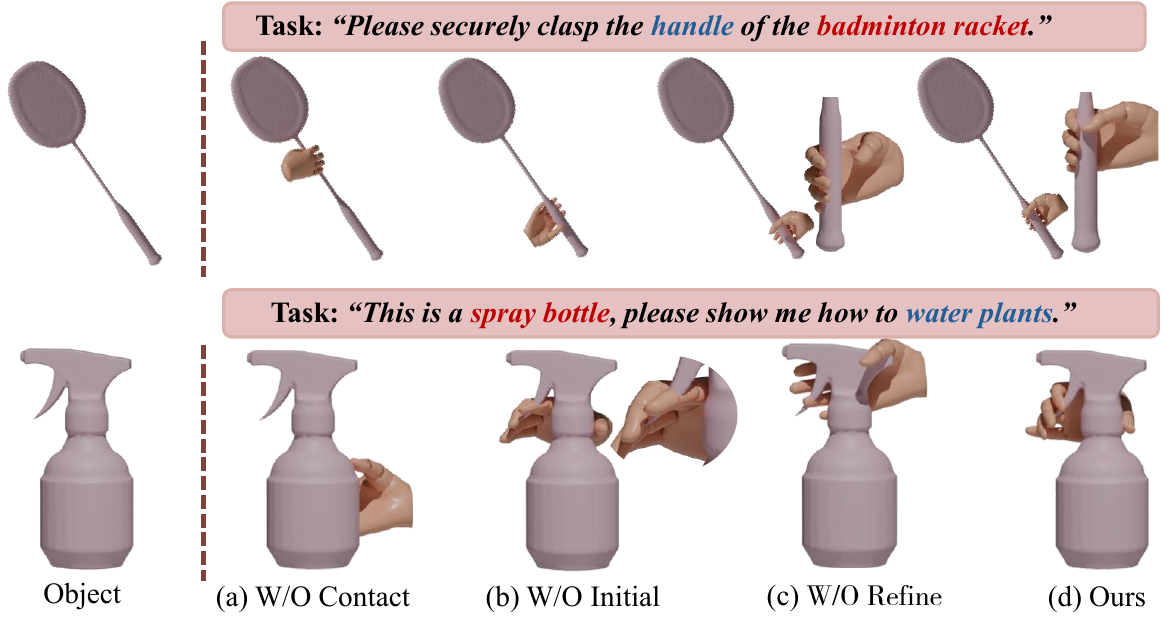}
   \caption{Qualitative evaluation of ablation study, illustrating the effects of contact inference (a), initial pose reasoning (b), and refinement (c).}
   \label{fig: ablation-results}
   \vspace{-18pt}
\end{figure}
\subsection{Ablation Study}

We perform ablation studies to examine the contribution of our design toward synthesizing semantic-aligned dexterous grasps. Qualitative and quantitative evaluations are reported in Fig.~\ref{fig: ablation-results} and Tab.~\ref{tab: ablation_results}.

\textbf{What is the effect of contact inference?} To investigate the role of contact inference, we remove the part- and point-level contact reasoning while keeping the inference of initial hand position, rotation, and grasp type based on the object center, as \emph{W/O Contact}. As shown in the first row of Tab.~\ref{tab: ablation_results}, removing contact inference slightly improves simulation stability, which is likely due to the hand near the object center. However, it results in a minor decrease in penetration control and a substantial drop in semantic alignment. Contact inference is indispensable for guiding grasps to task-relevant regions; without it, the hand often converges on incorrect areas, producing geometrically feasible yet semantically invalid grasps, as shown in Fig.~\ref{fig: ablation-results} (a). 

\textbf{Is contact inference alone sufficient for grasp semantic alignment?}
To evaluate the impact of initial grasp pose reasoning, we remove the zero-shot initial grasp pose inference stages, including hand position, rotation, and grasp type. The hand is instead initialized at the center of the part-level contact region $M_{hand}^{3d}$ and a random global rotation with joints in a natural extended configuration, as \emph{W/O Initial.}. Besides, while the target contact information $\{ M_{hand}^{3d}, P_{func}^{3d} \}$ is still provided as refinement guidance. In Tab.~\ref{tab: ablation_results}, although contact information is still provided for refinement, performance worsened in all metrics. As shown in the third column of Fig.~\ref{fig: ablation-results}, the hands exhibit unreasonable orientation and position. For example, in the badminton racket case, even when the hand contacts the correct region, its inverted orientation is inconsistent with human usage. In the spray bottle case, incorrect initialization causes severe penetration. In contrast, leveraging MLLMs for initial grasp reasoning provides strong guidance for semantic alignment and physical plausibility.

\textbf{Are initial grasp poses without refinement sufficiently accurate?}
We further analyze the effectiveness of our contact-guided grasp refinement by comparing optimized grasps with the unrefined initial grasp poses. This experiment clarifies whether this stage is necessary or if initial grasps already meet the semantic requirements and physical plausibility. As shown in Tab.~\ref{fig: ablation-results} (semantic metrics) and Fig.~\ref{fig: ablation-results} (c), the initial poses already capture semantic requirements, indicating that our prompt-based multi-stage reasoning component effectively bridges task semantics, object affordances, and grasp pose. However, results without refinement perform poorly in penetration and contact. In contrast, applying refinement not only ensures physical feasibility but also substantially improves semantic alignment.

\subsection{Real-World Experiments}

\begin{figure}[!t]
   \includegraphics[width=0.98\linewidth]{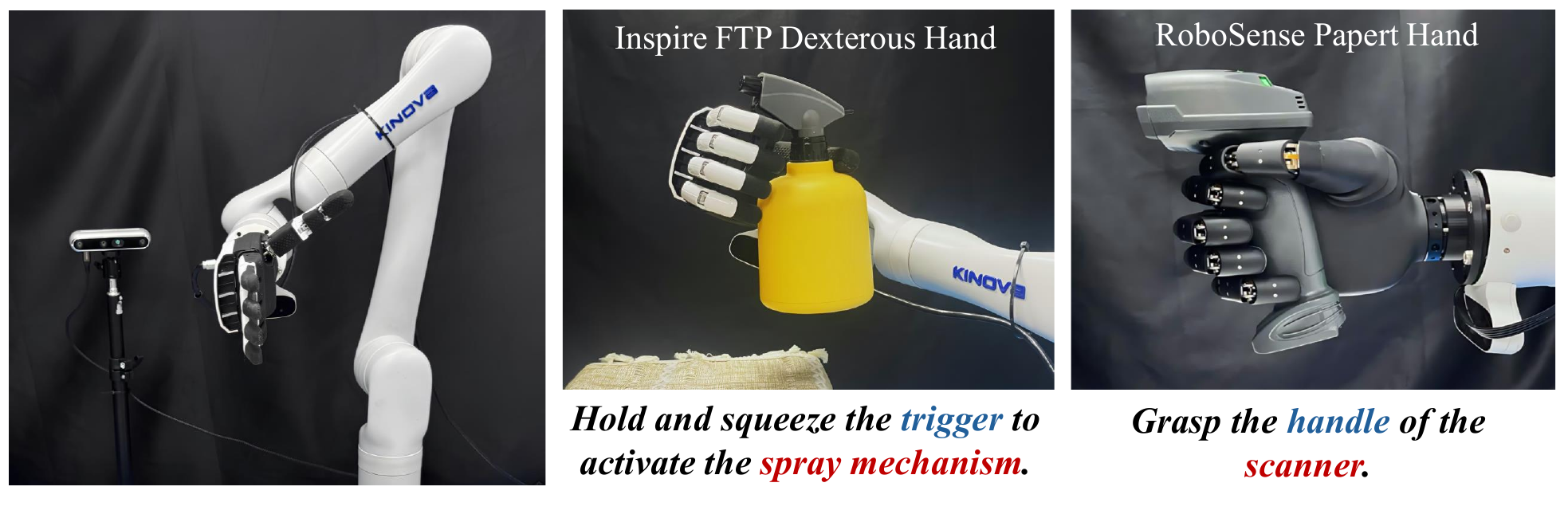}
   \caption{Qualitative results of Real-world.}
   \label{fig:real-world}
   \vspace{-18pt}
\end{figure} 

We evaluate our method on a real robotic platform consisting of a Kinova Gen3 arm equipped with an Inspire FTP dexterous hand and a RealSense D455 RGB-D camera; we also employ a RoboSense Papert Hand for further evaluation. Since our framework requires full object point clouds, we reconstruct a digital twin of the manipulated object from a single-view image. Specifically, we first apply the SAM~\cite{kirillov2023segany} to obtain the object mask, followed by 3D reconstruction~\cite{zhang2024clay} to generating realistic digital assets. The reconstructed digital twin is further aligned with the observed scene through real-to-sim adaptation, enabling accurate 6D pose estimation with FoundationPose~\cite{wen2024foundationpose}. To execute the grasps, we employ a finger keypoint-based retargeting method to transfer poses from the MANO model to the robotic hand. As illustrated in Fig.~\ref{fig:real-world}, this pipeline allows our method to remain compatible on real-world grasping tasks.

\section{CONCLUSION AND LIMITATION}

We presented ZeroDexGrasp, a zero-shot task-oriented dexterous grasp synthesis framework that combines MLLM-based semantic reasoning with contact-guided refinement. By bridging task semantics, object affordances, and grasp configurations. Experiments on diverse object categories and tasks demonstrate strong semantic alignment, physical plausibility, and real-world applicability.
However, our framework relies on MLLMs whose instability and hallucinations may lead to reasoning errors, which can affect grasp quality. We believe that with the rapid advancement of MLLMs, our approach holds great potential. Future work will focus on enhancing robustness through stronger spatial grounding and feedback mechanisms.


{
    \small
    \bibliographystyle{IEEEtran}
    \bibliography{root.bib}
}

\end{document}